\documentclass[one-column]{revtex4-1}
\usepackage{graphicx}

\begin{document}

\title{Reservoir-size dependent learning in analogue neural networks}

\author{Xavier Porte}
\email{javier.porte@femto-st.fr}
\affiliation{D\'{e}partement d'Optique P. M. Duffieux, Institut FEMTO-ST,  Universit\'e Bourgogne-Franche-Comt\'e CNRS UMR 6174, Besan\c{c}on, France.}

\author{Louis Andreoli}
\affiliation{D\'{e}partement d'Optique P. M. Duffieux, Institut FEMTO-ST,  Universit\'e Bourgogne-Franche-Comt\'e CNRS UMR 6174, Besan\c{c}on, France.}

\author{Maxime Jacquot}
\affiliation{D\'{e}partement d'Optique P. M. Duffieux, Institut FEMTO-ST,  Universit\'e Bourgogne-Franche-Comt\'e CNRS UMR 6174, Besan\c{c}on, France.}

\author{Laurent Larger}
\affiliation{D\'{e}partement d'Optique P. M. Duffieux, Institut FEMTO-ST,  Universit\'e Bourgogne-Franche-Comt\'e CNRS UMR 6174, Besan\c{c}on, France.}

\author{Daniel Brunner}
\affiliation{D\'{e}partement d'Optique P. M. Duffieux, Institut FEMTO-ST,  Universit\'e Bourgogne-Franche-Comt\'e CNRS UMR 6174, Besan\c{c}on, France.}


\begin{abstract}
The implementation of artificial neural networks in hardware substrates is a major interdisciplinary enterprise.
Well suited candidates for physical implementations must combine nonlinear neurons with dedicated and efficient hardware solutions for both connectivity and training. 
Reservoir computing addresses the problems related with the network connectivity and training in an elegant and efficient way. 
However, important questions regarding impact of reservoir size and learning routines on the convergence-speed during learning remain unaddressed. 
Here, we study in detail the learning process of a recently demonstrated photonic neural network based on a reservoir. 
We use a greedy algorithm to train our neural network for the task of chaotic signals prediction and analyze the learning-error landscape. 
Our results unveil fundamental properties of the system's optimization hyperspace. 
Particularly, we determine the convergence speed of learning as a function of reservoir size and find exceptional, close to linear scaling. 
This linear dependence, together with our parallel diffractive coupling, represent optimal scaling conditions for our photonic neural network scheme. 

\end{abstract}

\maketitle   

\section{\label{sec:Introduction}Introduction}
Nowadays neural networks (NNs) still remain extensively emulated by traditional computers, which posts important challenges in terms of parallelization, energy efficiency and overall computing speed. 
Ultimately, full hardware integration of NNs, where nonlinear nodes, network connectivity and optimization through learning are implemented via dedicated functionalities of the substrate is desirable. 
Optics-based solutions like optoelectronic~\cite{Paquot2012, Larger2012} or all-photonic~\cite{Duport2012, Brunner2013, VanDerSande2017} neural networks are of particular interest because they can avoid parallelization bottlenecks. 

In the context of hardware implementation, reservoir computing (RC) appeared as a particularly well suited approach to train and operate NNs~\cite{Maass2002, Jaeger2004a}. 
The convenience of RC originates in a training that is restricted to optimization of the readout weights, leaving the input as well as the internal connectivity between neurons unaffected.
However, in physically implemented NNs the training itself is conditioned by hardware structure. 
Here, one of the fundamental questions is how an error landscape is explored by a given learning algorithm when applied to a hardware NN. 
Therefore, understanding the topology of the cost function and its potential convexity or presence of local minima is of major importance. 

We experimentally implement RC in an optoelectronic analogue NN and train it to predict chaotic time series via a greedy algorithm, analogously to~\cite{Bueno2018a}. 
We study in detail the error-landscape, which we also refer as cost-function, differentiating those features caused by topology from those originated by noise. 
The mapped landscape is rich in features, on average follows an exponential topology and contains numerous local minima with comparable good-performance. 
We demonstrate that by using our greedy algorithm learning converges systematically.
Moreover, we address the particular question of how the NN rate of convergence and the prediction error depend on the network size. 
This is the first time that the fundamental characteristics of greedy learning are explored in a noisy physically implemented NN. 

\section{\label{sec:NeuralNetwork}Neural network concept and training}
Our optoelectronic NN is composed of up to 961 neurons whose state is encoded in the pixels of a spatial-light modulator (SLM). 
The neurons are connected among themselves via diffractive optical coupling, which is inherently parallel and scalable~\cite{photorescomp}. 

\subsection{\label{sec:ExpSetup}Experimental setup}
\begin{figure}
	\includegraphics[width=0.8\textwidth]{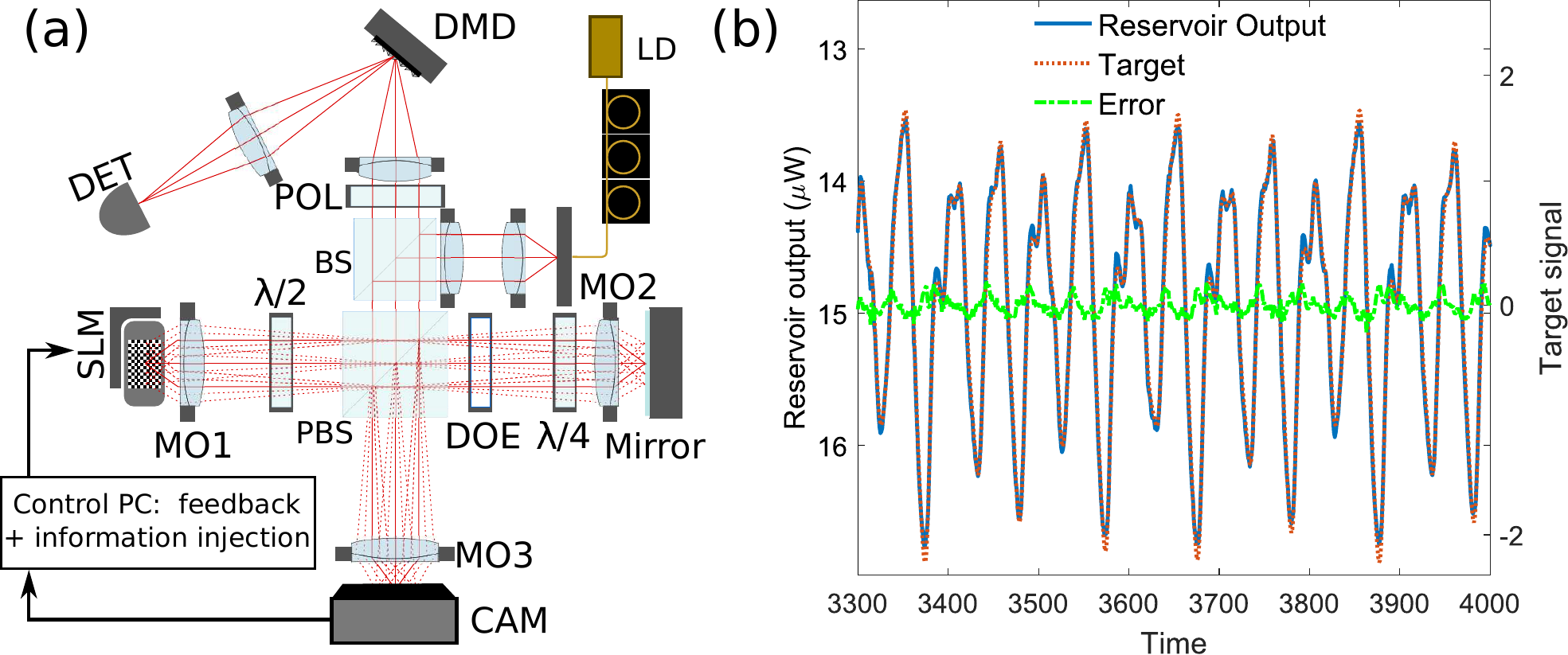}
	\caption{(a) Schematic of our recurrent neural network. (b) NN performance for the Mackey-Glass chaotic time series prediction task: the target chaotic time series, the reservoir’s output and the prediction error are depicted in orange, blue and green, respectively.} \label{fig1}
\end{figure}

The experimental implementation is schematically illustrated in Fig. \ref{fig1}(a). 
A laser diode of intensity $|E_{i}^{0}|^{2}$ illuminates the SLM, where the neurons' states $x_{i}$ are encoded. 
The SLM is imaged on the camera (CAM) after passing through the polarizing beam splitter (PBS) and twice through the diffractive optical element (DOE). 
The information detected by the camera is used to drive the SLM, realizing the NN recurrent connectivity. 
The dynamical evolution of the recurrent NN is given by	
\begin{equation}\label{eq:x(n+1)}		
x_{i}(n+1)= \alpha |E_{i}^{0}|^{2} cos^{2} \Bigg[ \beta \cdot \alpha \left| \sum_{j}^N W_{i,j}^{DOE} E_{j}(n)\right|^{2}+\gamma W_{i}^{inj} u(n+1)+\theta_{i} \Bigg],
\end{equation}	
where $E_j(n)$ is the optical electric field for each neuron, $\beta$ is the feedback gain, $\gamma$ is the input injection gain, $\alpha$ is an empirical normalization parameter, and $\theta_i$ are the phase offsets for each node. 
After optimization, operational parameters [$\beta, \gamma, \alpha, \theta_i$] are kept constant. 
The control PC reads the camera output and sets the new state of the SLM following Eq.~\ref{eq:x(n+1)}. 
Recurrency is stablished by previous neuron state $E_{j}(n)$, the external information is $u(n+1)$, $W^{DOE}$ is the recurrent neurons' internal coupling and $W^{inj}$ is the information injection matrix with random, independent and uniformly distributed weights between 0 and 1.

Following the RC principle, the NN training is restricted only to the modification of the readout weights. 
For that, the neurons of the recurrent NN, i.e. the SLM pixels, are imaged on a digital micro-mirror display (DMD) and focused on the surface of a photodiode.
The output of the neural network, as measured by the photodiode is
\begin{equation}\label{eq:Yout}
y^{out}(n+1)\propto\left|\sum_{i}^N W_{i}^{DMD} (E_{i}^{0}-E_{i}(n+1))\right|^{2},
\end{equation}
where $W^{DMD}_{i=1...N}$ are the optically implemented readout weights. 
The DMD mirrors can be flipped only between two positions, $\pm 12^{\circ}$. 
Thus, the readout weights are strictly Boolean and physically correspond to the orientation of the mirrors towards or away from the photodiode. 
By choosing which mirrors are directed towards the detector, we choose the set of active neurons that contribute to computing. 
Once trained, the DMD is turned into a passive device, operating without bandwidth limitation or energy consumption~\cite{Bueno2018a}. 

\subsection{\label{sec:greedyLearning}Training with a greedy algorithm}
Learning in our system optimizes the Boolean readout weights $W^{DMD}_{i=1...N,k}$ during successive learning epochs $k$ such that the output gradually approximates the desired response. 

Our greedy learning algorithm explores the error-landscape by favoring the selection of readout weights not tested yet.	
The vector $W^{select}_{k}$ is calculated at each iteration, giving a new value $W^{select}_{k}=rand(N)\cdot W^{bias}$ to each readout weight position. 
Here, $W^{bias}$ is a vector randomly initialized at the epoch $k=1$ and the function $rand(N)$ creates N random numbers uniformly distributed between 0 and 1. 
At every learning epoch, the algorithm chooses a new DMD position $l_{k}$ as the position of the maximum value of $W^{select}$, i.e. $l_{k} = max(W^{select}_k)$, and then changes its Boolean readout weight $W^{DMD}_{l_{k},k+1}=\neg \ W^{DMD}_{l_{k},k}$. 

For each epoch $k$ the mean square error (NMSE) $\epsilon_k$ is calculated. 
The error is defined as function of the normalized output $\tilde{y}^{out}(n+1)$ and the target signal $\mathcal{T}$, both normalized by the standard deviation and subtracted their offset: 
\begin{equation}\label{eq:epsilon}
	\epsilon_k = \frac{1}{T} \sum_{n=1}^{T}\left(\mathcal{T}(n+1)-\tilde{y} ^ {out} _ {k}(n+1)\right)^{2},
\end{equation}
where $T$ is the length of the chaotic time series used for training. 
We train for one-step-ahead prediction, and the target signal $\mathcal{T}(n + 1)$ is the injection signal one step ahead $u(n + 2)$. 
The calculation of the reward $r(k)$ is based on the performance evolution in comparison to the previous epoch as 
\begin{equation}\label{eq:r(k)}
	r(k) = 
	\left\{
	\begin{array}{r c l}  
	1 \quad \textrm{if}\ \epsilon_k < \epsilon_{k-1}\\
	0 \quad \textrm{if} \ \epsilon_k \ge \epsilon_{k-1}.
	\end{array}
	\right.
\end{equation}
The modified configuration of the DMD is kept depending on reward $r(k)$:
\begin{equation}\label{eq:WDMDmodifReward}
	W^{DMD}_{l_k,k} = r(k)W^{DMD}_{l_k,k}+(1-r(k))W^{DMD}_{l_k,k-1}. 
\end{equation}
If performance has not improved with respect to previous configuration, the DMD mirror is flipped back. 

In order to favor the selection of readout weights not yet tested, we implement 
\begin{equation}\label{eq:Wbias}
	W^{bias}=\frac{1}{N}+W^{bias}, \qquad W^{bias}_{l_k}=0,
\end{equation}
where the values of $W^{bias}$ are increased by $1/N$ at each epoch, setting afterwards the value assigned to the current $l_k$ position to zero.
Consequently, the bias for a previously modified weight increases linearly, approaching unity after $k=N$ learning iterations. 

Figure~\ref{fig1}(b) shows an example of the NN performance after training. 
The task is prediction of Mackey-Glass chaotic time series. 
The reservoir's output $y(n)$ is accurately predicting the next step of the chaotic time series $u(n+1)$. 

\section{\label{sec:ResultsLearning}Results on learning and error landscape}
Two hundred points of the chaotic Mackey-Glass sequence are used as training signal, and the same sequence is repeated for every learning epoch $k$.  
At each epoch, the greedy learning maximally modifies the value of a single readout-weight entry, hence varies the system's position within the error-landscape position by distance one. 
This is the maximal Hamming distance associated to every learning step.
An optimization path is a descent trajectory from the readout weight's starting configuration towards a minimum.

The optimized and system parameters are $\beta=0.8$, $\gamma=0.25$, $\theta_0=0.44\pi$, $\theta_0+\Delta\theta_0=0.95\pi$, $\mu=0.45$, $N=961$. 
The normalization parameter $\alpha$ represents necessary attenuation such the camera is not saturated. 

\subsection{\label{subsec:LearningAndTopology}Learning and topology}
The descending trend of learning for this system was already introduced in \cite{Bueno2018a}.
Here we want to explore the systematic characteristics of greedy learning. 
Since the optimization path in the error-landscape is randomized and subject to experimental noise, we study the statistical variability of learning by measuring twenty different learning curves with identical network parameters.
In order to restrict our findings to the properties of our learning-routine, we start all measurements at an identical position in the error-landscape, i.e. with an identical initial $W^{DMD}_{i,1}$.

\begin{figure}
	\centering	
	\includegraphics[width=0.6\textwidth]{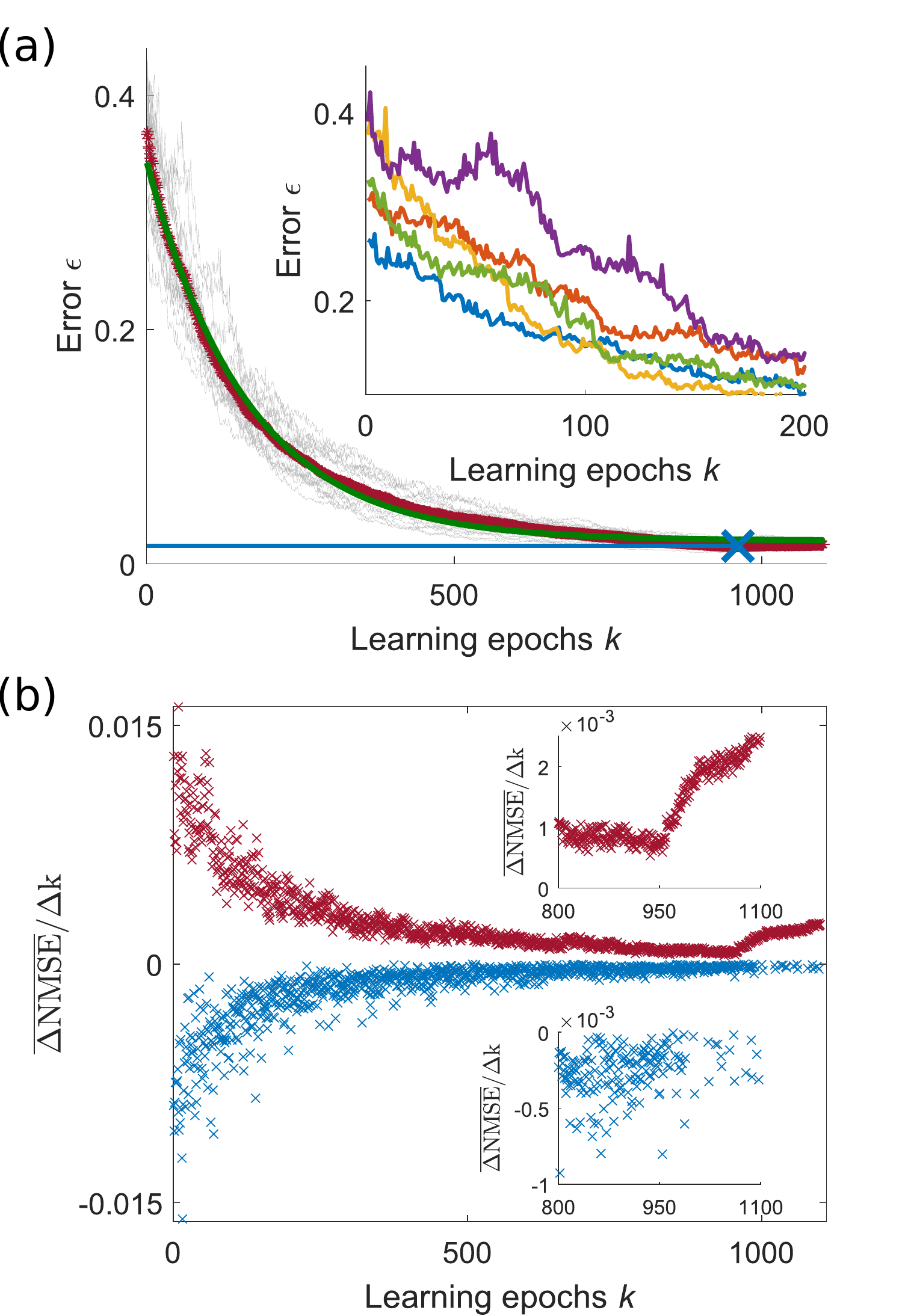}
	\caption{(a)Learning performance for the Mackey-Glass chaotic time series prediction. 
	Red asterisks are the average of the background 20 gray curves and the green line is its exponential fit. 
	The inset illustrates the strong initial local variability depicting the first 200 epochs for five exemplary learning curves. 
	(b) Average of positive (red) and negative (blue) error gradients of the 20 learning curves displayed in panel (a). 
	Insets: Zooms of the two curves around the epoch $k=950$. 
	}
	\label{fig2}
\end{figure}

The results are depicted in Fig.~\ref{fig2}(a). 
Twenty individual learning curves are presented in gray and their average is plotted in red. 
We can observe that all the curves converge towards a common minimum, after which all curves will slightly increase. 
The average value of this minimum is $\overline{\epsilon}_{k_{opt}} = (14.2\pm1.5)\cdot 10^{-3}$.
The green line is the exponential fit to the average of the 20 learning curves.  
The blue line illustrates the system's testing error, which we determined using a set of 9000 data-points not used during the training sequence. 
We observe that the testing error $\epsilon_{test} = 15\cdot 10^{-3}$ matches excellently with the learning error. 
From this result, we can conclude that no over-fitting is present, which we attribute to the role of noise in our analogue experimental NN. 

As shown in more detail in the inset of Fig. \ref{fig2}(a), individual learning curves follow different trajectories, ranging from a rather smooth descent to paths including steep jumps.
The large variability among the first learning epoch can be attributed to experimental system's noise because the initial DMD configuration is identical for all curves. 
However, the local variability during the learning process decreases and can therefore be related to particular topological properties of the various error-landscape explorations rather than to noise. 

In order to further study the optimization paths' topology and the global characteristics of  descent trajectories, we calculate the gradient of the average descent. 
At each learning epoch the relative change in local error is $\delta \epsilon / \delta k = \epsilon^{min} - \epsilon_{k}$, where $\epsilon^{min}$ is the previously smallest error achieved by the system. 
We define $\delta \epsilon ^+ / \delta k$ ($\delta \epsilon ^- / \delta k$) which contains all positive (negative) $\delta \epsilon /\delta k$. 
At each learning epoch we calculated the average $\delta \epsilon^{+} / \delta k$ and $\delta\epsilon^{-} / \delta k$, which correspond to the average positive and negative gradients of the error landscape at each learning epoch $k$. 
Data for the average positive (negative) gradients are shown as red (blue) in Fig.~\ref{fig2}(b).

Both, red and blue curves in Fig.~\ref{fig2}(b) exponentially decrease, which one could expect given that they represent the derivative of the exponentially decaying learning curves. 
This means that, on average, the error landscape curvature follows a decreasing exponential. 
Moreover, qualitatively different behaviour can be observed between both gradients when concentrating on the last learning epochs, cf. insets in Fig.~\ref{fig2}(b). 
While the negative gradients follow a noisy convergence towards zero, the positive gradients experience a sudden change of trend after the learning epoch $k \simeq 950$.
This iteration corresponds to the average epoch when the learning curves reach their minimum.
We therefore consider the NN learning process as composed of two parts: first an optimization path where the error and its gradient decrease, reaching a minimum in the error-landscape and becoming trapped. 
There is then a sharp increase in positive gradients, while at the same time negative gradients drop below the noise level.
From the clear trends of the positive gradients, we conclude that the optimization path around the minimum is not noise limited but defined by the topology of the error landscape.
In fact, due to the rule Eq.~\ref{eq:Wbias}, the probability of one readout weight to be modified again increase linearly in $1/N$. 
Therefore, after N epochs, the selection of position $l_k$ statistically repeats the selection sequence carried out during the first part of the optimization. 
Once trapped in a minimum, the sequence of tested dimensions reveal an inverse organization of the error gradient.
Consequently, the dimensions that have contributed strongest in reducing the error when optimized, lead now to least degradation of the performance.

\subsection{\label{subsec:LearningScalability}Learning scalability}	
We now focus on an interesting characteristic that emerges from Fig. \ref{fig2}(a), where we found that the optimization epochs until convergence to best performance are similar for all learning curves. 
The 20 different curves converge on average in $\sim 961$ epochs. 
Thus, the average number of iterations required to optimize the NN is of the order of the number of neurons. 

\begin{figure}
	\centering
	\includegraphics[width=0.6\textwidth]{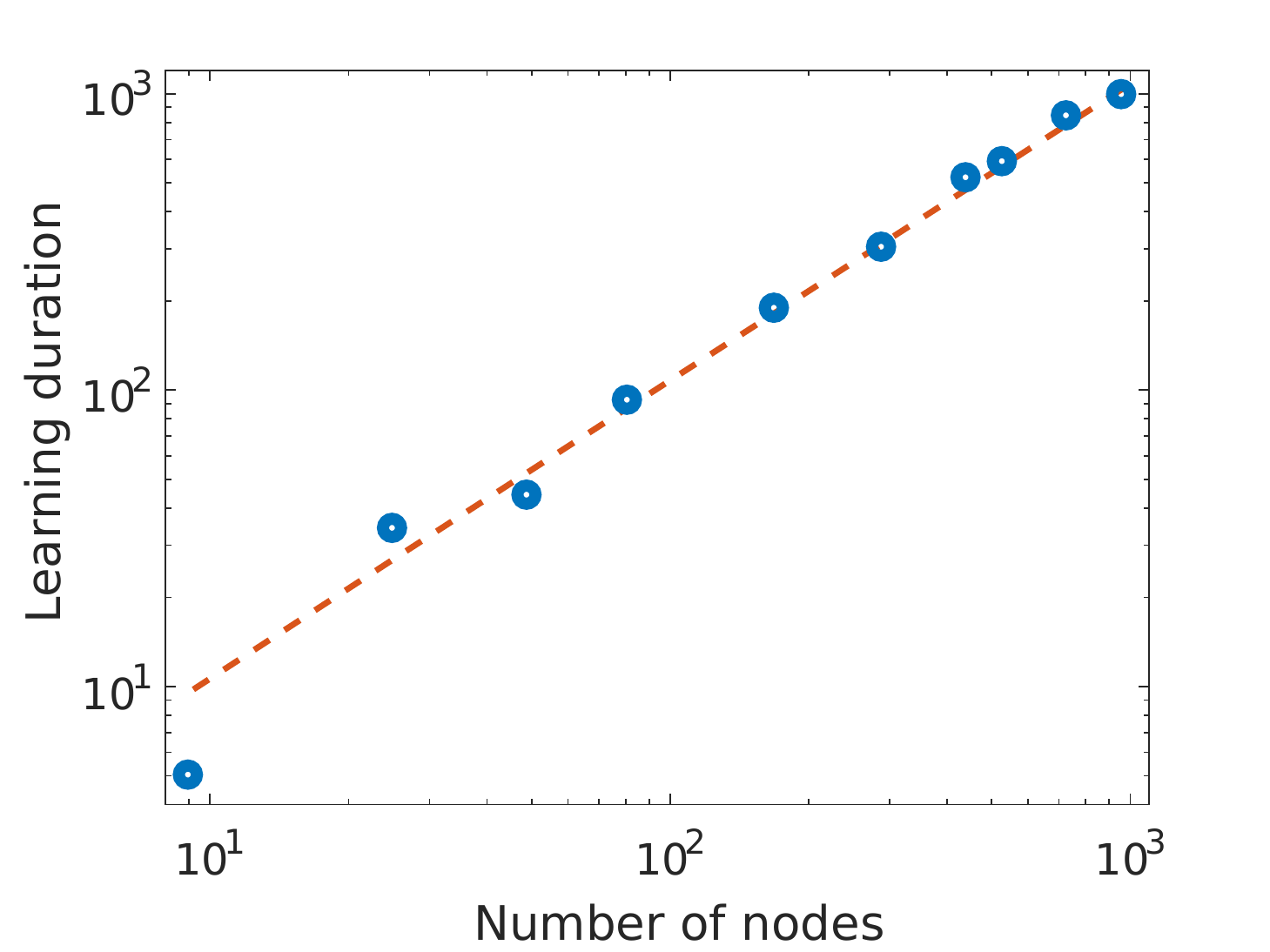}
	\caption{Scaling of the optimal learning epoch’s in function of the number of neurons. 
		Red dashed line shoes the polynomial fit to the data, obtaining a coefficient of 1.08 that indicates close-to-linear scaling.
	} 
	\label{fig3}
\end{figure}

We test now if this particular ratio maintains when modifying the size of the NN. 
Figure~\ref{fig3} shows the results of the optimal learning epoch for different network sizes ranging from 9 to 961 nodes.
Impressively, the experimental results (blue circles) have an almost perfect linear distribution over three orders of magnitude. 
The slope of the linear fit in logarithmic scale is 1.08.
Crucially, the prediction performance continuously improves for that larger NNs, hence optimization of all nodes is relevant also for the largest system.  
The 9 neurons network performs $\sim 50$ worse than the network with 961 neurons. 

\section{\label{sec:Conclusions}Conclusions}	
Our work addresses fundamental questions about the size-dependent performance of analogue NNs and about the topology of their learning-error landscape. 

We have investigated the features of greedy learning in analogue NNs. 
We have shown that applying our one-Boolean-step exploration algorithm, learning systematically converges towards similar minima. 
Nevertheless, different learning curves do not follow the same paths but topologically distinct trajectories. 
This suggests that all the optimization paths are ultimately trapped in distinct local minima with comparable prediction performance.

We have also experimentally demonstrated that the duration and effectiveness of the training are clearly correlated to the NN size. 
In particular, the number of epochs required for optimal learning scales almost perfect linearly with the NN size. 
This is a crucial finding that combines with the inherent parallel nature of diffractive coupling to boost the scalability our photonic NN approach. 

\subsection*{Acknowledgements}
This work has been supported by the EUR EIPHI program (Contract No. ANR-17-EURE-0002), by the BiPhoProc ANR project (No. ANR-14-OHRI-0002-02), by the Volkswagen Foundation NeuroQNet project and the ENERGETIC project of Bourgogne Franche-Comt\'{e}.
X.P. has received funding from the European Union’s Horizon 2020 research and innovation programme under the Marie Sklodowska-Curie grant agreement No. 713694 (MULTIPLY).

%
%
\bibliography{references}

\end{document}